\documentclass[11pt,a4paper]{article}
\usepackage[hyperref]{emnlp2018}
\usepackage{times}
\usepackage{latexsym}
\usepackage{graphicx}
\usepackage{booktabs}
\usepackage{enumitem,multirow}
\usepackage{multicol}

\usepackage{url}

\usepackage{todonotes}   

\aclfinalcopy 

\title{A strong baseline for question relevancy ranking}

\author{Ana V. Gonz\'alez-Gardu\~no  and
  Isabelle Augenstein and
  Anders S{\o}gaard \\
  Dpt.~of Computer Science\\
   University of Copenhagen\\
  $\{${\tt ana}$\mid${\tt augenstein}$\mid${\tt soegaard}$\}${\tt @di.ku.dk} \\}

\date{}

\begin{document}
\maketitle
\begin{abstract}
The best systems at the SemEval-16 and SemEval-17 community question answering shared tasks -- a task that amounts to {\em question relevancy ranking} -- involve complex pipelines and manual feature engineering. Despite this, many of these still fail at beating the IR baseline, i.e., the rankings provided by Google's search engine. We present a strong baseline for question relevancy ranking by training a simple multi-task feed forward network on a bag of 14 distance measures for the input question pair. This baseline model, which is fast to train and uses only language-independent features, outperforms the best shared task systems on the task of retrieving relevant previously asked questions.

\end{abstract}

\section{Introduction}

Community question-answer fora are great resources, collecting answers to frequently and less-frequently asked questions on specific topics, but these are often not moderated and contain many irrelevant answers. Community Question Answering (CQA), cast as a question relevancy ranking problem, was the topic of two shared tasks at SemEval 2016-17. This is a non-trivial retrieval task, typically evaluated using mean average precision (MAP). We present a strong baseline for this task, on par with or surpassing state-of-the-art systems.

The English subtasks of the SemEval CQA \cite{nakov2015semeval,nakov2017semeval} consist of Question-Question Similarity, Question-Comment Similarity, and Question-External Comment Similarity. 
In this study, we focus on the core subtask of Question-Question similarity, defined as follows: Given a question, rank other relevant questions by their relevancy to that question. This proved to be a difficult task in both SemEval-16 and SemEval-17 as it is the one with the least amount of data available. The baseline was the ranking retrieved by performing a Google search, which proved to be a strong baseline beating a large portion of the systems submitted. 

\paragraph{Contribution} Our baseline is a simple multi-task feed-forward neural network taking distance measures between pairs of questions as input. We use a question-answer dataset as auxiliary task; but we also experiment with datasets for pairwise classification tasks such as natural language inference and fake news detection. This simple, easy-to-train model is on par or better than state-of-the-art systems for question relevancy ranking. We also show that this simple model outperforms a more complex model based on recurrent neural networks.

\section{Our Model}

We present a simple baseline model for question relevancy ranking.\footnote{Code available at http://anavaleriagonzalez/FAQ\_rank.} It is a deep feed-forward network with a hidden layer that is shared with an auxiliary task model. The input to the network is extremely simple and consists of five distance measures of the input question-question pair. \S2.1 discusses these distance measures, and how they relate. \S2.2 introduces the multi-task learning architecture that we propose. 

\subsection{Features}
We use four similarity metrics and three sentence representations (averaged word embeddings, binary unigram vectors, and trigram vectors). The {\bf cosine distance} between the sentence representations of query x and query y is $$\frac{\sum_i\mathbf{x}_i\mathbf{y}_i}{\sqrt[]{\sum_i\mathbf{x}^2}+\sqrt[]{\sum_i\mathbf{y}^2}}$$ The {\bf Manhattan distance} is $$\sum_i|\mathbf{x}_i-\mathbf{y}_i|$$ The {\bf Bhattacharya distance} is $$-\ln(\sum_i\sqrt[]{\mathbf{x}_i\mathbf{y}_i})$$ and is a measure of divergence, and the {\bf Euclidean distance} is $$\sqrt[]{\sum_i(\mathbf{x}_i-\mathbf{y}_i)^2}$$ Note that the squared Euclidean distance is proportional to cosine distance and Manhattan distance. The Bhattacharya and Jaccard metrics, on the other hand, are sensitive to the number of types in the input (the $\ell_1$ norm of the vector encodings). So, for example, {\em only} the cosine, Euclidean, and Manhattan distances will be the same for 


\begin{footnotesize}$$\mathbf{x}=\langle 1,1,0,0,1,0,1,1,0,1\rangle,\mathbf{y}=\langle0,0,1,0,1,0,0,0,1,1\rangle$$\end{footnotesize} 


and 

\vspace*{-4mm}

\begin{footnotesize}$$\mathbf{x}=\langle 0,0,0,0,0,1,0,0,1,1\rangle,\mathbf{y}=\langle1,1,1,1,0,0,0,0,0,1\rangle$$\end{footnotesize}

\vspace*{-5mm}

The {\bf Jaccard index} is the only metric that can only be applied to two of our representations, unigrams and $n$-grams: It is defined over $m$-dimensional {\em binary} (indicator) vectors and therefore not applicable to averaged embeddings. It is defined as $$\frac{\mathbf{x}\cdot\mathbf{y}}{m}$$ We represent each query pair by these 14 numerical features.

\begin{figure}[h]
\includegraphics[width=7cm]{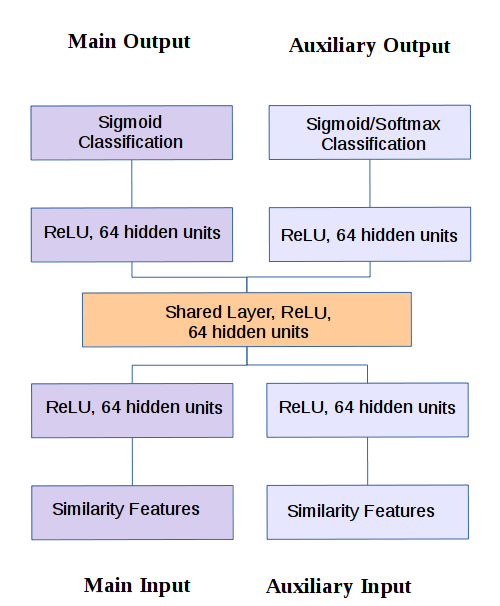}
\label{MLP}
\caption{The architecture of the multi task learning MLP }
\end{figure}

\subsection{MTL Architecture}
Our architecture is a simple feed-forward, multi-task learning (MTL) architecture. Our architecture is presented in Figure 1 and is a Multi-Layer Perceptron (MLP) that takes a pair of sequences as input. The sequences can be sampled from the main task or the auxiliary task. The MLP has one shared hidden layer, a task-specific hidden layer and, finally, a task-specific classification layer for each output. The hyper-parameters, after doing grid search, optimizing performance on the validation data, are given in Figure~\ref{hp}. 

\begin{figure*}
\begin{center}
\footnotesize
\begin{tabular}{|c|c|c|}
 \hline
 Hyperparameter  & Best Value & Tested Values\\ 
 \hline
 Num. of Epochs &  100 & 10, 20, 40, 60, 80, 100\\ 
 Batch Size &  100 &10, 50, 100\\ 
 Learning rate & 0.001 & 0.001, 0.01, 0.1, 0.2, 0.3 \\ 
 Momentum &  0.9 & 0.0, 0.2, 0.4, 0.6, 0.8, 0.9\\ 
 Dropout &  0.02  & 0.0, 0.1, 0.2, 0.3, 0.4, 0.5, 0.6, 0.7, 0.8, 0.9\\ 
 \hline
 
\end{tabular}
\caption{\label{hp}Hyper-parameters}
\end{center}
\end{figure*}

\subsection{LSTM baseline}
We compare our MLP ranker to a bidirectional LSTM \cite{hochreiter1997long} model. It takes two sequences inputs: sequence 1 and sequence 2, and a stack of three bidirectional LSTM layers, which encode sequence 1 and sequence 2, respectively. The outputs are then concatenated, to enable representing the differences between the two sequences. Instead of relying only on this presentation \cite{bowman2015large,conf/emnlp/AugensteinRVB16}, we also concatenate our distance features and feed everything into our MLP ranker described above.


\begin{table*}[ht!]
\centering
\begin{footnotesize}
\begin{tabular}{lllll|llll}\\
\toprule
& \multicolumn{3}{c}{\textsc{EN-SemEval-16}} & \multicolumn{3}{c}{\textsc{EN-SemEval-17}}\\
\midrule
Model & \textsc{MAP} & \textsc{MRR} & \textsc{AvgRec} & \textsc{Acc} & \textsc{MAP} & \textsc{MRR} & \textsc{AvgRec} & \textsc{Acc}  \\
\midrule
 STL-MLP (our baseline)  &  76.21 &  84.17 &  88.85  &   77.86 &   47.12 &   51.44 &  81.95&  63.30 \\
STL-BiLSTM-SIM  &  73.42 &  82.02 &  87.93 &   76.35&   46.85 &   51.20 &  81.00& 68.64 \\

IR baseline   &  74.75 &  83.79  &  88.30 & - &  41.85 &   46.42 &   77.59 &   - \\
Random baseline   &  46.98 &  50.96  &  67.92 &   - &   29.81   &  33.02 & 62.65 &  - \\
SemEval-Best   &  76.70&  83.02  &  90.31 &   76.57 &  47.22 &  50.07 &  82.60 & 52.39 \\
\midrule
\multicolumn{9}{c}{MTL - English Only - QA Aux}\\
\midrule
MTL-MLP \  & \textbf{77.61**} &  \textbf{84.29**} &   \textbf{90.33**} & 78.99** &    47.66** &   \textbf{52.41**} & 82.37** &  71.70**  \\
MTL-BiLSTM-SIM &  71.25 &  79.54 &  86.52&  75.26 & 43.58 &  49.16 &  79.36 & 50.40\\
\midrule
\multicolumn{9}{c}{MTL - FNC Aux}\\
\midrule
MTL-MLP \  & 76.83 &  84.05 &  90.29 & \textbf{79.30} &  47.33**&  49.80** & \textbf{83.63**} &  73.06** \\
MTL-BiLSTM-SIM &  71.70 &  80.83 & 87.35 &   73.29 & 48.15* &  51.90* & 81.59* & 64.00* \\
\midrule
\multicolumn{9}{c}{MTL - NLI Aux}\\
\midrule
MTL-MLP \  & 77.06** & 83.98** &  89.97**  & 78.57** &  48.00** &  51.56** & 82.91** & 70.11** \\
MTL-BiLSTM-SIM &  75.22** &  83.02** &  89.14**&   77.71 & \textbf{48.56*}  & 52.05* & 82.22* & 71.36*\\
\midrule
\multicolumn{9}{c}{MTL - ALL Aux}\\
\midrule
MTL-MLP \  & 75.90** & 84.17** & 89.59**  &78.51** &  47.45** &  51.12** & 82.90** &  \textbf{73.10**} \\
\bottomrule

\end{tabular}
\end{footnotesize}
\caption{The results show that learning pairwise classification tasks simultaneously with the main task leads to improvements over the baselines and the best SemEval systems. We show results for three auxiliary tasks, Question-Comment relevancy prediction, Fake News detection and Natural Language Inference. The asterisks for the MTL results represent the significance of the improvements over the STL systems with ** representing a p-value of $< 0.01 $ and * representing a p-value between 0.01 and 0.05 }
\label{tab:results}
\end{table*}

\section{Datasets}

For our experiments, we use data from SemEval shared tasks, but we also take advantage of potential synergies with other existing datasets for classification of sentence pairs. 
Below we present the datasets used for our main and auxiliary tasks. We provide some summary statistics for each dataset in Table \ref{data_stats}.

\paragraph{SemEval 2016 and 2017} As our main dataset we use the queries from SemEval's subtask B which consists of an original query and 10 possibly related queries. As an auxiliary task, we use the data from subtask A, which is a question-related comment ranking task. 

\paragraph{Natural Language Inference}

Natural Language Inference (NLI), consists in predicting {\sc entailment}, {\sc contradiction} or {\sc neutral}, given a hypothesis and a premise. 
We use the MNLI dataset as opposed to the SNLI data \cite{bowman2015large,nangia2017repeval}, since it contains different genres. Our model is not built to be a strong NLI system; we use the similarity between premise and hypothesis as a weak signal to improve the generalization on our main task.

\paragraph{Fake News Challenge}
The Fake News Challenge\footnote{http://www.fakenewschallenge.org/} (FNC) was introduced to combat misleading and false information online. This task has been used before in a multi-task setting as a way to utilize general information about pairwise relations \cite{augenstein2018multi}. 
Formally, the FNC task consists in, given a headline and the body of text which can be from the same news article or not, classify the stance of the body of text relative to what is claimed in the headline. There are four labels:
\begin{itemize}[noitemsep]
\item \textsc{Agrees}: The body of the article is in agreement with the headline
\item \textsc{Disagrees}: The body of the article is in disagreement with the headline
\item \textsc{Discusses}: The body of the article does not take a position
\item \textsc{Unrelated}: the body of the article discusses a different topic
\end{itemize}

We include fake news detection as a weak auxiliary signal that can lead to better generalization of our question-question ranking model.

\begin{figure}
\begin{center}
\begin{tabular}{c|ccc}
 \toprule
 
Dataset & Train &Dev& Test\\
 \midrule 
 SemEval 16 &  2000 & 500 &700\\ 
 SemEval 17 &  - & - &880\\ 
 FNC & 49k & -  & - \\ 
 MNLI & 433k & - & - \\ 
 \bottomrule
 
\end{tabular}
\caption{\label{data_stats}Size of datasets used for the experiments. Here we present the full size of FNC and MultiNLI training sets, however for our MTL experiments we used a random sample of the same size of the train, test and dev sets of the SemEval data as auxiliary data. The SemEval 17 shared task uses the same train and dev set as SemEval 16}
\end{center}
\end{figure}

\subsection{Evaluation} 

We evaluate our performance on the main task of question relevancy ranking using the official SemEval-2017 Task 3 evaluation scripts \cite{nakov2017semeval}. The scripts provide a variety of metrics; however, in accordance with the shared task, we report Mean Average Precision (MAP) (the official metric for the SemEval 2016 and 2017 shared tasks); Mean Reciprocal Rank (MRR), which has being thoroughly used for IR and QA; Average Recall; and, finally, the accuracy of predicting relevant documents.

\section{Results}

The results from our experiments are shown in Table \ref{tab:results}. We present the official metric from the SemEval task, as well as other common metrics. For the SemEval-16 data, our multitask MLP architecture with a question-answer auxiliary task performed best on all metrics, except accuracy, where the multi-task MLP using all auxiliary tasks performed best. We outperform the winning systems of both the SemEval 2016 and 2017 campaigns. In addition, our improvements from single-task to multi-task are significant ($p<0.01$). We also outperform the official IR baseline used in the SemEval 2016 and 2017 shared tasks. We discuss the STL-LSTM-SIM results in \S5. Furthermore, in Table \ref{tab:group-vectors}, we show the performance of our models when training on feature combinations, while in Table \ref{tab:ablation2},  we present an ablation test where we remove one feature at a time. 

\begin{figure}[h]
\includegraphics[width=8.5cm,height=7cm]{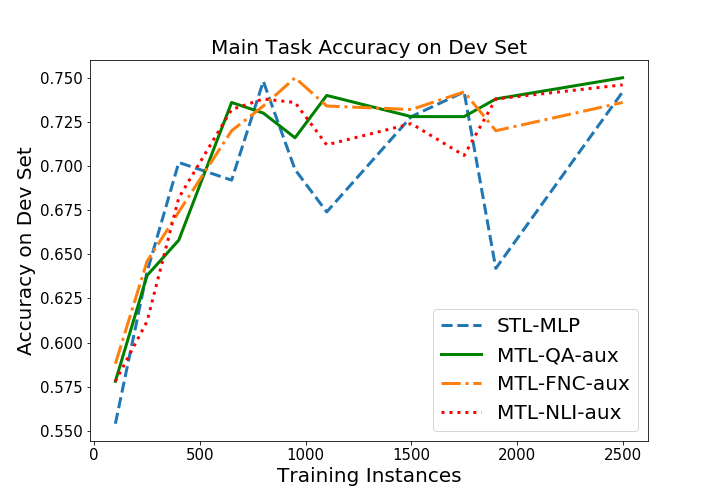}
\label{Learningcurve}
\caption{Learning curves for single- and multi-task learning. All MTL models outperform the STL model with very few training samples, and learning curves are more stable for MTL models than for STL.}
\end{figure}

\paragraph{Learning curve} In Figure \ref{Learningcurve}, we also present our learning curves for the development set when incrementally increasing the training set size. We observe that when using an auxiliary task, the learning is more stable across training set size. 

\begin{table}[h!]
\centering
\begin{footnotesize}
\begin{tabular}{l|ll|ll}
\toprule
 \multicolumn{1}{c}{} & \multicolumn{2}{c}{\textsc{STL}} & \multicolumn{2}{c}{\textsc{MTL}}  \\
\midrule
 \textsc{Feature Set} &  \textsc{MAP} &  \textsc{ACC} &  \textsc{MAP} &  \textsc{ACC} \\
 \midrule

Unigrams & 72.20 & 74.21 & \textbf{73.37} & \textbf{74.9}  \\
MeanEmb & \textbf{67.00} & 71.28 & 65.39 &  \textbf{72.29}\\
Trigrams & \textbf{72.8} & 77.8 & 72.00 & \textbf{78.2} \\
Unigrams+Ngrams & 73.62 & 77.00 & \textbf{74.01} & \textbf{78.60} \\
Unigrams+MeanEmb & 62.06 &72.60 & \textbf{70.68} & \textbf{74.14}  \\
Emb+Trigrams & 66.00 & 75.01 & \textbf{72.25} & \textbf{77.48} \\

\bottomrule

\end{tabular}
\end{footnotesize}
\caption{Performance on development set of SemEval-16 when training {\em only} on certain feature combinations. MTL uses QA as auxiliary data.}
\label{tab:group-vectors}
\end{table}

\begin{table}[h!]
\centering
\begin{footnotesize}
\begin{tabular}{ll|ll}

\toprule
 \multicolumn{2}{c}{\textsc{Feature Removed}} &  \textsc{MAP} &  \textsc{ACC}  \\
 \midrule
\multirow{3}{*}{\sc Cosine}&unigram&69.25 & {\bf 74.99}\\
&trigram& 69.93& 76.28 \\
&embedding& 69.11 & 76.40\\
\midrule
\multirow{3}{*}{\sc Manhattan}&unigram& 70.33& 76.71\\
&trigram& 69.29 &76.28 \\
&embedding& 66.90 & 75.28 \\
\midrule
\multirow{3}{*}{\sc Bhattacharya}&unigram& 70.83& 75.85\\
&trigram& 71.14 & 77.50 \\
&embedding& 71.72& 77.28 \\
\midrule
\multirow{3}{*}{\sc Euclidean}&unigram& 71.43& 76.57\\
&trigram& {\bf 65.55}& 76.14 \\
&embedding& 70.60 & 75.57 \\
\midrule
\multirow{2}{*}{\sc Jaccard}&unigram& 67.41 & 75.70\\
&trigram& 69.27 & 75.98 \\
\bottomrule
\end{tabular}
\end{footnotesize}
\caption{We perform an ablation test, where we remove one feature at a time and report performance on development data of our single-task baseline. We observe that our baseline suffers most from removing the Euclidean distance over trigrams and the cosine similarity over unigrams. Note also that the Jaccard index over unigrams seems to carry a strong signal, albeit a very simple feature.} 
\label{tab:ablation2}
\end{table}

\section{Discussion}

For the SemEval shared tasks on CQA, several authors used complex recurrent and convolutional neural network architectures \cite{severyn2015learning,barron2016convkn}. For example, \citeauthor{barron2016convkn} used a convolutional neural network in combination with feature vectors representing lexical, syntactic, and semantic similarity as well as tree kernels. Their performance was slightly lower than the best system (SemEval-Best for 2016 in Table \ref{tab:results}). The best system used lexical and semantic similarity measures in combination with a ranking model based on support vector machines (SVMs) \cite{filice2016kelp,franco2016uh}. Both systems are harder to implement and train than the model we propose here. 
For SemEval-17, \citet{franco2016uh}, the winning team used distributed representations of words, knowledge graphs and frames from FrameNet \cite{baker1998berkeley} as some of their features, and used SVMs for ranking.

For a more direct comparison, we also train a more expressive model than the simple MTL-based model we propose. This architecture is based on bi-directional LSTMs \cite{hochreiter1997long}. For this model, we input sequences of embedded words (using pre-trained word embeddings) from each query into independent BiLSTM blocks and output a vector representation for each query. We then concatenate the vector representations with the similarity features from our MTL model and feed it into a dense layer and a classification layer. This way we can evaluate the usefulness of the flexible, expressive LSTM network directly (as our MTL model becomes an ablation instance of the full, more complex architecture). We use the same dropout regularization and SGD values as for the MLP. Tuning all parameters on the development data, we do not manage to outperform our proposed model, however. See lines MTL-LSTM-SIM in Table~1 for results.

\section{Conclusion}
We show that simple feature engineering, combined with an auxiliary task and a simple feedforward neural architecture is appropriate for a small dataset and manages to beat the baseline and the best performing systems for the Semeval task of question relevancy ranking. We observe that introducing pairwise classification tasks leads to significant improvements in performance and a more stable model. Overall, our simple model introduces a new strong baseline which is particularly useful when there is a lack of labeled data.  

\section*{Acknowledgments}
The first author of this paper is funded by a BotXO PhD Award;\footnote{http://www.botxo.co/} the last author by an ERC Starting Grant. We gratefully acknowledge the support of the NVIDIA Corporation with the donation of the Titan Xp GPU used for this research.

\bibliography{emnlp2018}
\bibliographystyle{acl_natbib_nourl}

\appendix

\end{document}